\documentclass[format=acmsmall, review=false, screen=true, manuscript=true]{acmart}

\usepackage{booktabs} 
\usepackage{graphicx}

\usepackage{amssymb}

\usepackage{caption}

\usepackage[inline]{enumitem} 

\makeatletter
\newcommand{\labitem}[2]{%
	\def\@itemlabel{\textbf{#1}}
	\item
	\def\@currentlabel{#1}\label{#2}}
\makeatother

\makeatletter
\newcommand{\headingitem}[1]{%
	\vspace{0.3cm}
	\def\@itemlabel{\textbf{#1}}
	\item
	\def\@currentlabel{#1}
	\addtocounter{enumi}{-1}
}
\makeatother

\usepackage{csvsimple}
\usepackage{multirow}
\usepackage{lipsum}

\usepackage{eurosym}
\usepackage{siunitx}

\DeclareSIUnit\eur{\officialeuro}
\DeclareSIUnit\M{M}
\DeclareSIUnit\k{k}

\def\sym#1{\ifmmode^{#1}\else\(^{#1}\)\fi}

\usepackage{setspace} 
\usepackage{csquotes}
\usepackage{amsmath}

\usepackage{xspace}
\newcommand\ie{i.\,e.\xspace}
\newcommand\eg{e.\,g.\xspace}

\newcommand\etc{etc.\xspace}

\newcommand\cf{cf.\xspace}

\makeatletter
\let\copy@theorem@headerfont=\theorem@headerfont
\newcommand{\my@theorem@headerfont}{%
	\boldmath\copy@theorem@headerfont\unboldmath
}
\let\theorem@headerfont=\my@theorem@headerfont
\makeatother

\usepackage[%
sort&compress
]{cleveref}
\usepackage{url}

\DeclareMathOperator*{\argmin}{arg\,min}

\usepackage[ruled]{algorithm2e} 

\SetAlFnt{\small}
\SetAlCapFnt{\small}
\SetAlCapNameFnt{\small}
\SetAlCapHSkip{0pt}
\IncMargin{-\parindent}

\usepackage{colortbl}
\usepackage{tabularx}
\usepackage{booktabs}
\usepackage{collcell}

\usepackage{ragged2e}

\newcommand{\PreserveBackslash}[1]{\let\temp=\\#1\let\\=\temp}
\newcolumntype{v}[1]{>{\PreserveBackslash\RaggedRight\hspace{0pt}}p{#1}}

\usepackage{adjustbox}
\newcolumntype{Q}[2]{%
	>{\adjustbox{angle=#1,lap=\width-(#2)}\bgroup}%
	l%
	<{\egroup}%
}

\makeatletter
\DeclareRobustCommand\onedot{\futurelet\@let@token\@onedot}
\def\@onedot{\ifx\@let@token.\else.\null\fi\xspace}

\def\eg{e.g\onedot} 
\def\ie{i.e\onedot} 
\def\cf{c.f\onedot} 
\def\etc{etc\onedot}

\makeatother

\newcommand{\mcellt}[2][c]{%
	\begin{tabular}[t]{@{}#1@{}}#2\end{tabular}}

\usepackage[
colorinlistoftodos,
textsize=footnotesize,
]{todonotes}

\makeatletter
\renewcommand{\fps@figure}{htb}         
\renewcommand{\fps@table}{htb}         
\makeatother 

\usepackage{float}
\usepackage{rotating}

\setcopyright{None}

\begin{document}

\title[Domain Customization for Question Answering]{Putting Question-Answering Systems into Practice:\\ Transfer Learning for Efficient Domain Customization}

\author{Bernhard Kratzwald}
\orcid{}
\affiliation{%
  \institution{ETH Zurich}
  \streetaddress{Weinbergstrasse 56}
  \city{Zurich}
  \postcode{8006}
  \country{Switzerland}}
\email{bkratzwald@ethz.ch}
\author{Stefan Feuerriegel}
\orcid{}
\affiliation{%
	\institution{ETH Zurich}
	\streetaddress{Weinbergstrasse 56}
	\city{Zurich}
	\postcode{8006}
	\country{Switzerland}}
\email{sfeuerriegel@ethz.ch}

\begin{abstract}
	Traditional information retrieval (such as that offered by web search engines) impedes users with information overload from extensive result pages and the need to manually locate the desired information therein. Conversely, question-answering systems change how humans interact with information systems: users can now ask specific questions and obtain a tailored answer -- both conveniently in natural language. Despite obvious benefits, their use is often limited to an academic context, largely because of expensive domain customizations, which means that the performance in domain-specific applications often fails to meet expectations. This paper proposes cost-efficient remedies: (i) we leverage metadata through a filtering mechanism, which increases the precision of document retrieval, and (ii) we develop a novel fuse-and-oversample approach for transfer learning in order to improve the performance of answer extraction. Here knowledge is inductively transferred from a related, yet different, tasks to the domain-specific application, while accounting for potential differences in the sample sizes across both tasks. The resulting performance is demonstrated with actual use cases from a finance company and the film industry, where fewer than 400 question-answer pairs had to be annotated in order to yield significant performance gains. As a direct implication to management, this presents a promising path to better leveraging of knowledge stored in information systems.
\end{abstract}



\maketitle

\renewcommand{\shortauthors}{B. Kratzwald and S. Feuerriegel}
\section{Introduction}


Question-answering (Q\&A) systems redefine interactions with management information systems~\cite{Lim.2013} by changing how humans seek and retrieve information. This technology replaces classical information retrieval with natural conversations~\cite{Simmons.1965}. In traditional information retrieval, users query information systems with keywords in order to retrieve a (ranked) list of matching documents; yet a second step is necessary in which the user needs to extract the answer from a particular document~\cite{Belkin.1993,Chau.2008}. Conversely, Q\&A systems render it possible for users to directly phrase their question in natural language \emph{and} also retrieve the answer in natural language. Formally, such systems specify a mapping $(q, D) \mapsto a$ in order to search an answer $a$ for a question $q$ from a collection of documents $D = [d_1, d_2, \ldots ]$. Underlying this approach is often a two-step process in which the Q\&A system first identifies the relevant document $d_q \in D$ within the corpus and subsequently infers the correct answer $a \in d_q$ from that document \cite[\cf][]{Moldovan.2003}.


Question-answering systems add several benefits to human-computer interfaces: first, question answering is known to come more naturally to humans than keyword search, especially for those who are not digital natives~\cite[\cf][]{Vodanovich.2010}. As a result, question answering presents a path for information systems that can greatly contribute to the ease of use~\cite{Radev.2002} and even user acceptance rates~\cite{Giboney.2015,Schumaker.2007}. Second, Q\&A systems promise to accelerate the search process, as users directly obtain the correct answer to their questions~\cite{Roussinov.2007}. In practice, this obviates a large amount of manual reading necessary to identify the relevant document and to locate the right piece of information within one. Third, question answering circumvents the need for computer screens, as it can even be incorporated into simple electronic devices (such as wearables or Amazon's Echo). 


One of the most prominent Q\&A systems is IBM Watson~\cite{Ferrucci.2012}, known for its 2011 win in the game show \textquote{Jeopardy}. IBM Watson has since grown beyond question answering, now serving as umbrella term that includes further components from business intelligence. The actual Q\&A functionality is still in use, predominantly for providing healthcare decision support based on clinical literature.\footnote{IBM Watson for Oncology. \url{https://www.ibm.com/watson/health/oncology-and-genomics/oncology/}, accessed January~8, 2018.} Further research efforts in the field of question answering have led to systems targeting applications, for instance, from medicine~\cite[\eg][]{Cao.2011}, education~\cite[\eg][]{Cao.2004} and IT security~\cite{Roussinov.2007}. However, the aforementioned works are highly specialized and have all been tailored to the requirements of each individual use case.


Besides the aforementioned implementations, question-answering technology has found very little adoption in actual information systems and especially knowledge management systems. From a user point of view, the performance of current Q\&A systems in real-world settings is often limited and thus diminishes user satisfaction. The predominant reason for this is that each application requires cost-intensive customizations, which are rarely undertaken by practitioners with the necessary care. Individual customization can apply to, \eg, domain-specific knowledge, terminology and slang. Hitherto, such customizations demanded manually-designed linguistic rules \cite{Kaisser.2004} or, in the context of machine learning, extensive datasets with hand-crafted labels \cite[\cf][]{Ling.2017}. Conversely, our work proposes an alternative strategy based on transfer learning. Here the idea is an inductive transfer of knowledge from a general, open-domain application to the domain-specific use case~\cite[\cf][]{Pan.2010}. This approach is highly cost-efficient as it merely requires a small set of a few hundred labeled question-answer pairs in order to fine-tune the machine learning classifiers to domain-specific applications.


We conduct a systematic case study in which we tailor a Q\&A system to two different use cases, one from the financial domain and one from the film industry. Our experiments are based on a generic Q\&A system that allows us to run extensive experiments across different implementations. We have found that conventional Q\&A systems can answer only up to one out of \num{3.4} questions correctly in the sense that the proposed answer exactly matches the desired word sequence (\ie not a sub-sequence and no redundant words). Conversely, our system achieves significant performance increases as it bolsters the correctness to one out of \num{2.0} questions. This is achieved by levers that target the two components inside content-based Q\&A systems. A filtering mechanisms incorporates metadata of documents inside information retrieval. Despite the fact that metadata is commonly used in common in knowledge bases, we are not aware of prior uses cases within content-based systems for neural question answering. We further improve the answer extraction component by proposing a novel variant of transfer learning for this purpose, which we call \emph{fuse-and-oversample}. It is key for domain customization and accounts for a considerable increase in accuracy by \SI{8.1}{\percent} to \SI{17.0}{\percent}. In fact, both approaches even benefit from one another and, when implemented together, improve performance further.


The remainder of this paper is organized as follows. \Cref{sec:bg} reviews common research streams in the field of question-answering systems with a focus on the challenges that arise with domain customizations. We then develop our strategies for domain customization -- namely, metadata filtering and transfer learning -- in \Cref{sec:methods}. The resulting methodology is evaluated in \Cref{sec:results}, demonstrating the superior performance over common baselines. Based on these findings, \Cref{sec:discussion} concludes with implications for the use of Q\&A technology in management information systems. 

\section{Background}
\label{sec:bg}

Recent research on question answering can be divided into two main paradigms according to how these systems reason the response: namely, (i)~ontology-based question answering that first maps documents onto entities in order to operate on this alternative representation and (ii)~content-based systems that draw upon raw textual input. 

Deep learning is still a nascent tool in information systems research and, since we make extensive use of it, we point to further references for the interested reader. A general-purpose introduction is given in \cite{Goodfellow.2016}, while the work in \cite{Kraus.2018} studies the added value to firm operations. A detailed overview of different architectures can be found in \cite{Schmidhuber.2015}.  

\subsection{Ontology-Based Q\&A Systems}



One approach to question answering is to draw upon ontology-based representations. For this purpose, the Q\&A system first transforms both questions and documents into ontologies, which are then used to reason the answer. The ontological representation commonly consists of semantic triples in the form of \texttt{<subject, predicate, object>}. In some cases, the representation can further be extended by, for instance, relational information or unstructured data~\cite{Xu.2016}. The deductive abilities of this approach have made ontology-based systems especially prevalent in relation to (semi-)structured data such as large-scale knowledge graphs from the Semantic Web~\cite[\eg][]{Unger.2012,Berant.2013,Ferrandez.2009,Lopez.2007}. 


In general, ontology-based Q\&A systems entail several drawbacks that are inherent to the internal representation. On the one hand, the initial projection onto ontologies often results in a loss of information \cite{Vallet.2005}. On the other hand, the underlying ontology itself is often limited in its expressiveness to domain-specific entities \cite[\cf][]{Molla.2007} and, as a result, the performance of such systems is hampered when answering questions concerning previously-unseen entities. Here the conventional remedy is to manually encode extensive domain knowledge into the system \cite{Maedche.2001}, yet this imposes high upfront costs and thus impedes practical use cases. 

\subsection{Content-Based Q\&A Systems}
\label{sec:opendomainqa}


Content-based Q\&A systems operate on raw text, instead of the rather limited representation of ontologies~\cite[\eg][]{Harabagiu.2000, Radev.2002, Cao.2011}. For this reason, these systems commonly follow a two-stage approach~\cite{Jurafsky.2009}. In the first step, a module for information retrieval selects the relevant document $d_q$ from the corpus $D$ based on similarity scoring. Here the complete content of the original document is retained by using an appropriate mathematical representation (\ie tf-idf as commonly used in state-of-the-art systems). In the second step, the retrieved documents $d_q$ are further processed with the help of an answer extraction module that infers the actual response $a \in d_q$. The latter step frequently draws upon machine learning models in order to benefit from trainable parameters.


Content-based systems overcome several of the weaknesses of ontology-based approaches. First, the underlying similarity matching allows these systems to answer questions that involve out-of-domain knowledge (\ie unseen entities or relations in question). Second, content-based systems circumvent the need for manual rule engineering, as the underlying rules can be trained with machine learning. As a result, content-based Q\&A systems are often the preferred choice in practical settings. We later study how this type facilitates domain customization via transfer learning and how it benefits from advanced deep neural network architectures.

\subsubsection{Information Retrieval Module}
\label{sec:related-ir}


The first component filters for relevant documents based on the similarity between their content and the query~\cite{Jurafsky.2009}. For this purpose, it is convenient to treat documents as mathematical structures with a well-defined similarity measure. A straightforward approach is to transform documents into a vector space that represents documents and queries as sparse vectors of term or $n$-gram frequencies within a high-dimensional space \cite{Salton.1971}. The similarity between documents and queries can then be formalized as the proximity of their embedded vectors. 


In practice, raw term frequencies cannot account for the specificity of terms; that is, a term occurring in multiple documents carries less relevance. Therefore, plain term frequencies have been further weighted by the inverted document frequency, in order to give relevance to words that appear in fewer documents \cite{SparckJones.1972}. This tf-idf scheme incorporates an additional normalization on the basis of document length to facilitate comparison \cite{Singhal.1996}. Retrieval based on tf-idf weights has become the predominant choice in Q\&A systems~\cite[\eg][]{Buckley.1997, Harabagiu.2000} and has been shown to yield competitive results~\cite{Voorhees.2001}, as it is responsible for obtaining state-of-the-art results \cite{Chen.2017}.  

To overcome the limitations of bag-of-words features, state-of-the-art systems take the local ordering of words into account~\cite{Chen.2017}. This is achieved by calculating tf-idf vectors over $n$-grams rather than single terms. Since the number of $n$-grams grows exponentially with $n$, one usually utilizes feature hashing~\cite{Weinberger.2009} as a trade-off between performance and memory use.

\subsubsection{Answer Extraction Module}



The second stage derives the actual answer from the selected document. It usually includes separate steps that extract candidate answers and rank these in order to finally select the most promising candidate. We note that some authors also refer to this task as machine comprehension, predominantly when it is used in an isolated setting outside of Q\&A systems.


A straightforward approach builds upon the document extracted by the information retrieval module and then identifies candidate answers simply by selecting complete sentences~\cite{Richardson.2013}. More granular answers are commonly generated by extracting sub-sequences of words from the original document. These sub-sequences can either be formulated in a top-down process with the help of constituency trees~\cite[\cf][]{Radev.2002, Shen.2006, Rajpurkar.2016} or in a bottom-up fashion where $n$-grams are extracted from documents and subsequently combined to form longer, coherent answers~\cite[\cf][]{Brill.2002, Lin.2007}. 


A common way to rank candidates and decide upon the final answer is based on linguistic and especially syntactic information~\cite[\cf][]{Pasca.2005}. This helps in better matching the type of the information requested by the question (\ie time, location, \etc) with the actual response. For instance, the question \emph{\textquote{When did X begin operations?}} implies the search for a time and the syntactic structure of candidate answers should thus be fairly similar to expressions involving temporal order, such as \emph{\textquote{X began operations in Y}} or \emph{\textquote{In year Y, X began operations}}~\cite[\cf][]{Kaisser.2004}. In practice, the procedures of ranking and selection are computed via feature engineering along with machine learning classifiers~\cite{Echihabi.2008,Rajpurkar.2016,Ravichandran.2001}. We later follow the recent approach in \cite{Rajpurkar.2016} and utilize their open-source implementation of feature engineering and machine learning as one of our baselines.

Only recently, deep learning has been applied by \cite{Chen.2017, Wang.2017.b, Wang.2018} to the answer extraction module of Q\&A systems, where it outperforms traditional machine learning. In these works, recurrent neural networks iterate over the sequence of words in a document of arbitrary length in order to learn a lower-dimensional representation in their hidden layers and then predict the start and end position of the answer. As a result, this circumvents the need for hand-written rules, mixture of classifiers or schemes for answer ranking, but rather utilizes a single model capable of learning all steps end-to-end. Hence, we draw upon the so-called DrQA network from \cite{Chen.2017} as part of our experiments. 

Beyond that, we later experiment with further network architectures as part of a holistic comparison. Moreover, the machine learning classifiers inside answer extraction modules are known to require extensive datasets and we thus suggest transfer learning as a means of expediting domain customization.

\subsection{Transfer Learning}
\label{sec:bg_transfer_learning}

Due to the complexity of contemporary deep neural networks, one commonly requires extensive datasets with thousands of samples for training all their parameters in order to prevent overfitting and obtain a satisfactory performance~\cite{Rajpurkar.2016}. However, such large-scale datasets are extremely costly to acquire, especially for applications that require expert knowledge. A common way to overcome the limitations of small training corpora is transfer learning~\cite{Pan.2010}. 

The na\"ive approach for transfer learning is based on initialization and subsequent fine-tuning; that is, a model is first trained on a source task $S$ using a (usually extensive) dataset $D_S$. In a second step, the model weights are optimized based on the actual target task $T$ with (a limited or costly) dataset $D_T$. Prior studies show that this setting usually requires extensive parameter tuning to avoid over- or underfitting in the second training phase~\cite{Mou.2016}. 

A more complex approach for transfer learning is multi-task learning. In this setting, the model is simultaneously trained on a source and target task, that can be related \cite{Kratzwald.2018b} or equivalent \cite{Kraus.2017}. Since multi-task learning alternates between optimization steps on source and target task, it is not suitable for very small datasets as the one used in our study. 

As a remedy, we develop a novel \emph{fuse-and-oversample} variant of transfer learning, that presents a robust approach for inductive transfer towards small datasets. In fact, the implicit oversampling is a prerequisite in order to facilitate learning in our setting, where the dataset $D_S$ can comprise of thousands or millions of observations, while $D_T$ is limited to a few hundred.

\section{Methods and Materials}
\label{sec:methods}

\begin{figure}[b]
	\begin{center}
		\includegraphics[width=1\linewidth]{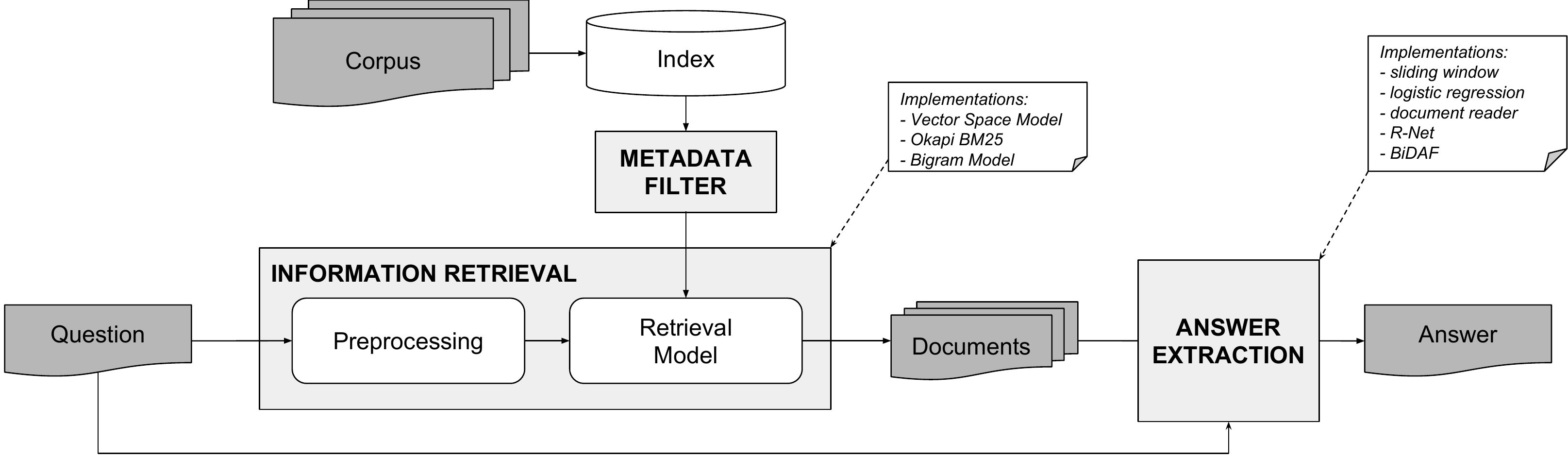}
	\end{center}
	\caption{Two-stage architecture of our generic content-based Q\&A system. The first stage draws upon functionality from the field of information retrieval in order to identify relevant materials, while the second stage generates the final answer. For both components we use different implementations to account for robustness in our analysis. The metadata filtering is placed upstream the information retrieval module.}
	\label{fig:qa_general}
\end{figure}

For this work, we designed a generic content-based Q\&A system as shown in \Cref{fig:qa_general}. Our generic architecture ensures that we can plug-in different modules for information retrieval and answer extraction throughout our experiments in order to demonstrate the general validity of our results. In detail, we are using up three state-of-the-art modules for information retrieval and five different modules for answer extraction. The implementation of these modules follows existing works and its description is thus summarized in \Cref{sec:appendix}.


The performance of question-answering techniques is commonly tested in highly artificial settings, \ie consisting of datasets that rarely match the characteristics of business settings. Based on our practical experience, we identify two important levers that help in tailoring Q\&A systems to specific applications, namely, (i)~an additional metadata filter (as shown in \Cref{fig:qa_general}), that restricts the scope of selected documents for answer extraction and (ii)~transfer learning as a tool for re-using knowledge from open-domain question-answering tasks. These mechanisms target different components of a Q\&A system: the metadata filter affects the behavior of the IR module, while transfer learning addresses the answer extraction. Both approaches are introduced in the following.

\subsection{Metadata Filter: Domain Customization for Information Retrieval}

The accumulated knowledge in content-based Q\&A systems is composed by its underlying corpus of documents. While in most applications, the system itself covers a wide area of knowledge, the individual documents address only a certain aspect of that knowledge. For instance, a corpus of financial documents contains knowledge about important events for many companies, while a single document might relate only to a change in the board of one specific company. Similarly, a single question can be answered by looking at a subset of relevant documents. For instance, the question \emph{"Who the current CEO of company X?"} can be answered by restricting the system to look at documents issued by company \emph{X}. Similarly, a user might ask \emph{"In what time span was John Sculley CEO?"} and restrict the answer to lie in documents issued by the entities \emph{Apple Inc.} or \emph{Pepsi-Cola}.

To determine this relevant subset of documents, we suggest to draw upon the metadata of documents. Metadata stores important information complementing the content, \eg timestamps, authors, topics or keywords. Metadata can even be domain-specific and, for example in the medical context, comprise of health record data that belongs to a specific patient, is issued by a certain hospital or doctor, and usually carries multiple timestamps. While knowledge-based Q\&A systems \cite[\cf][]{Pinto.2002} or community question-answering systems \cite[\cf][]{Bian.2008} rely heavily on metadata, this information has been so far ignored by content-based systems.

In our Q\&A system, we incorporate additional metadata by a filtering mechanism which is integrated upstream the information retrieval module. The metadata filter allows the user to restrict the set of relevant documents. This occurs prior to the question-document scoring step in the information retrieval module.

We implemented our system in a generic way, as it automatically displays input fields for every metadata field existing in the domain-specific corpus.  For categorical metadata (\ie list of companies, patient names, news categories, \etc), we display a drop-down field to select the desired metadata attribute or possibly multiple choices. For timestamps, and real-valued fields the user can select a lower and upper bound. This generic implementation allows for cost-efficient domain customization without further development costs.

\subsection{Domain Customization through Transfer Learning}

We develop two different approaches for transfer learning. The first approach is based on na\"ive initialization and fine-tuning as used in the literature (see \Cref{sec:bg_transfer_learning}). In short, the network is first initialized with an open-domain dataset consisting of more than 100,000 question-answer pairs. In a second step, we fine-tune the parameters of that model with a small domain-specific dataset. The inherent disadvantage of this approach is that the results are highly dependent on the chosen hyperparameters (\ie learning-rate, batch size, \etc), which need to be carefully tuned in each stage and thus introduce considerably more instability during training.

Fine-tuning is well known to behave unstable with tendencies to overfit \cite{Mou.2016,Siddhant.2018}. This effect is amplified when working with small datasets as they prohibit extensive hyperparameter tuning with regard to, \eg, learning rate, batch size. Since our objective is to propose a domain customization that requires fewer than thousands of samples, we expect initialization and fine-tuning to be very delicate. Therefore, we use it in our comparisons for reasons of completeness, but, in addition, develop a second approach for transfer learning which we name \emph{fuse-and-oversample}. This approach is based on re-training the entire model from scratch on a merged dataset. Hence, there is no direct need for users to change the original hyperparameters.

Our aim in applying transfer learning is to adjust the model towards the domain-specific language and terminology. By merging our domain-specific dataset with another large-scale dataset, we overcome the problem of having not sufficient training data. However, the resulting dataset is highly imbalanced and the fraction of training samples containing domain-specific language and terminology is well below one percent. To put more emphasis on the domain-specific samples, we borrow techniques from classification with imbalanced classes. Classification with imbalances classes is a similar but refers to a more simplistic problem, as the number of possible outcomes in classification is usually well below the number of training samples. In our case, we are predicting the actual answer (or, more precisely, a probability over it) within an arbitrary paragraph of text -- a problem fundamentally different from classification. 

Undersampling the large-scale dataset \cite[\eg][]{Liu.2009} would theoretically put more emphasis on the target domain but contradict our goal of having enough data-samples to train neural networks. Weighted losses \cite[\eg][]{Zong.2013} are often used to handle imbalances, but this concept is difficult to translate to our setting as we do not perform classification. Therefore, we decided to use an approach based on oversampling \cite[\eg][]{Japkowicz.2002}. By oversampling domain-specific question-answer pairs in a ratio of one to three, each epoch on the basic datasets accounts for three epochs of domain-specific fine-tuning, without the need to separate this phases or set different hyperparameters for them. This allows our fuse-and-oversample approach to perform an inductive knowledge transfer even when the domain-specific dataset is fairly small and consists only of a few hundred documents. In fact, our numerical experiments later demonstrate that the suggested combination of fusing and oversampling is key to obtain a successful knowledge transfer across datasets.

\section{Dataset}
\label{sec:dataset} 

We demonstrate our proposed methods for domain customization using an actual application of question answering from a business context, \ie where a Q\&A system answers question regarding firm developments based on news. We specifically decided upon this use case due to the fact that financial news presents an important source of information for decision-making in financial markets~\cite{Granados.2010}. Hence, this use case is of direct importance to a host of practitioners, including media and investors. Moreover, this setting presents a challenging undertaking, as financial news is known for its complex language and highly domain-specific terminology. 


Our dataset consists of financial news items (\ie so-called ad~hoc announcements) that were published by firms in English as part of regulatory reporting rules and were then disseminated through standardized channels. We proceeded with this dataset as follows: A subset of these news items was annotated and then split randomly into a training set (\SI{60}{\percent} of the samples), as well as a test set (\SI{40}{\percent}). As a result, we yield 63 documents with a total of 393 question-answer pairs for training. The test set consists of another 63 documents with 257 question-answer pairs, as well as 13,272 financial news items without annotations. This reflects the common nature of Q\&A systems that have to extract the relevant information oftentimes from thousands of different documents. Hence, this is necessary in order to obtain a realistic performance testbed for the overall system in which the information retrieval module is tested.

\Cref{tab:qa-samples} provides an illustrative set of question-answer pairs from our dataset.

\begin{table}[h]
	\centering
	\scriptsize
	{
		\begin{tabular}{p{7cm}p{3cm}p{3cm}}
			\toprule
			\multicolumn{1}{c}{\textbf{Document}} &\multicolumn{1}{c}{\textbf{Question}} &\multicolumn{1}{c}{\textbf{Answer}} \\
			\midrule
			\ldots Dialog Semiconductor PLC  (xetra: DLG), a provider of highly integrated power management, AC/DC power conversion, solid state lighting and Bluetooth(R) Smart wireless technology, today reports Q4 2015 IFRS revenue of \textbf{\$397 million}, at the upper end of the guidance range announced on 15 December 2015. \ldots &What is the level of Q4 2015 IFRS revenue for Dialog Semiconductor PLC?&\$397 million \\[0.3em]\\
			\ldots Dr. Stephan Rietiker, CEO of LifeWatch, stated: ``his clearance represents a
			significant technological milestone for LifeWatch and strengthens our position
			as an innovational leader in digital health. Furthermore, it allows us to
			commence our cardiac monitoring service in \textbf{Turkey} with a patch product
			offering.'' \ldots &Where does LifeWatch plan to start the cardiac monitoring service?& Turkey\\
			\bottomrule
		\end{tabular}
	}
	\caption{Two samples for question-answer pairs. The table shows the snippet of the news item, together with the location of the (shortest) ground-truth answer within it. }
	\label{tab:qa-samples}
\end{table} 


All documents were further subject to conventional preprocessing steps~\cite{Manning.1999}, namely, stopword removal and stemming. The former removes common words carrying no meaningful information, while the latter removes the inflectional form of words and reduces them to their word-stem. For instance, the words \emph{fished}, \emph{fishing} and \emph{fisher} would all be reduced to their common stem \emph{fish}.

Our application presents a variety of possibilities for implementing a metadata filtering, such as making selections by industry sector, firm name or the time the announcement was made. We found it most practical to filter for the firm name itself, since questions mostly relate to specific news articles. For instance, the question \emph{"What is the adjusted net sales growth at actual exchange rates in 2017?"} cannot be answered uniquely without defining a company. Hence, our experiments draw upon filtering by firm name. This also provides direct benefits in practical settings, as it saves the user from having to type such identifiers and thus improves the overall ease-of-use.


We further draw upon a second dataset during transfer learning, the prevalent Stanford Question and Answer (SQuAD) dataset~\cite{Rajpurkar.2016}. This dataset is common in Q\&A systems for general-purpose knowledge and is further known for its extensive size, as it contains a total of \num{107785} question-answer pairs. Hence, when merging the SQuAD and our domain-specific dataset, the latter only amounts to a small fraction of \SI{0.6}{\percent} of all samples. This explains the need for oversampling, such that the neural network is trained with the domain-specific question-answer pairs to a sufficient extent. 

\section{Results}
\label{sec:results}


In this section, we compare both strategies for customizing question-answering systems to business applications. Since each approach addresses a different module within the Q\&A system, we first evaluate the sensitivity of interactions with the corresponding component in an isolated manner. We finally evaluate the complete system as once. This demonstrates the effectiveness of the proposed strategies for domain customization separately and the added value when being combined.

\subsection{Metadata Filtering}
\label{sec:results_information_retrieval}


This section examines the performance of the information retrieval module, \ie isolated from the rest of the Q\&A pipeline. This allows us to study the interactions between the metadata filter and the precision of the document retrieval. We implemented a variety of approaches for document retrieval, namely, a vector-space model based on cosine similarity scoring~\cite{SparckJones.1972}, a probabilistic Okapi BM25 retrieval model~\cite{Robertson.2009}, and a tf-idf model based on hashed bigram counts~\cite{Chen.2017}. All models are described in \Cref{sec:information_retrieval}.


The information retrieval module is evaluated in terms of recall@$k$. This metric measures the ratio of how often the relevant document $d_q$ is within the top-$k$ ranked documents. More than one document can include the desired answer $a$ and we thus treat all documents with $a \in d$ as a potential match.  

\Cref{tab:results_information_retrieval} shows the numerical outcomes. 
We observe that the model based on hashed bigram counts generally outperforms the simple vector space model, as well as the Okapi BM25 model. This confirms the recent findings in \cite{Chen.2017,Wang.2018}, which can be attributed to the question-like input that generally contains more information than keyword-based queries. More importantly, exchanging the model results only in marginal performance changes. However, we observe a considerable change in performance when utilizing metadata information, where almost all models perform equally good. Here the recall@$1$ improves from \num{0.48} to \num{0.88}; \ie an increase of \num{0.4}. The increase become evident especially when returning one document opposed to five, since the corresponding relative improvement amounts to a \SI{83.3}{\percent} in terms recall@$1$ and only \SI{45.5}{\percent} for the recall@$5$.

The results demonstrate the immense quality of information stored in the metadata. With a simple filtering mechanism for the domain-specific attribute \emph{"firm name"} we can achieve a considerable gain in performance. The metadata filter reduces the number of potentially relevant documents for a query, yielding a probability of \num{0.46}, on average, of randomly choosing the right document for a given query. This highlights the potential of information stored in metadata for content-based Q\&A systems.

\begin{table}[t]
	\centering
	\small
	\begin{tabular}{l ccc}
		\bottomrule
		Approach & \multicolumn{1}{c}{Recall@$1$}     & \multicolumn{1}{c}{Recall@$3$}    & \multicolumn{1}{c}{Recall@$5$}      \\ 
		\midrule 
		\multicolumn{4}{l}{\emph{IR module without metadata filter}} \\
		\quad Vector Space Model      	 	 	  & 0.35           & 0.49          & 0.54            \\
		\quad Okapi BM25	&	0.41	&	0.52	&	0.55\\
		\quad Bigram Model    	 	  & \bfseries 0.48           & \bfseries 0.62          & \bfseries 0.66  	 \\
		\midrule
		\multicolumn{4}{l}{\emph{IR module with metadata filter}} \\
		\quad Baseline: random choice    	 	 & 0.46  	  & 0.76  	  & 0.88 	 \\
		\quad Vector Space Model  	 	 & 0.84           & 0.95          & 0.96 	 \\
		\quad Okapi BM25	& 0.87 &\bfseries 0.96 &\bfseries 0.98\\
		\quad Bigram Model  	 & \bfseries 0.88  	  & \bfseries 0.96  	  & \bfseries 0.98   	 \\
		\bottomrule
	\end{tabular}
	\caption{Comparison of how different variants of information retrieval affect the performance of this module. Here the average recall is measured when returning the top-$k$ documents (the best score for each choice of $k$ is highlighted in bold). As we can see, the performance changes only slightly when exchanging the retrieval model, yet the metadata filer corresponds to notable improvements.}
	\label{tab:results_information_retrieval}
\end{table}

\subsection{Answer Extraction Module}
\label{sec:results_answer_extraction}


This section studies the sensitivity of implementing domain customization within the answer extraction module. Accordingly, we specifically evaluate how transfer learning increases the accuracy of answer extraction and we compute the number of matches, given that the correct document is supplied, in order to assess the performance of this module in an isolated manner. That is, we specifically measure the performance in terms of locating the answer $a$ to a question $q$ in a given document $d_q$. We study three different neural network architectures, namely DrQA~\cite{Chen.2017}, BiDAF~\cite{Seo.2016} and R-Net~\cite{Wang.2017}. All three are described in the \Cref{sec:answer_extraction}


The results of our experiments are shown in \Cref{tab:results_answer_extraction}. Here we distinguish three approaches: (i)~the baseline without transfer learning for comparison, (ii)~the na{\"i}ve initialize and fine-tune approach to transfer learning whereby the networks are first trained based on the open-domain dataset before being subsequently fine-tuned to the domain-specific application, and (iii)~our fuse-and-oversample approach whereby we create a fused dataset such that the network is simultaneously trained on both open-domain and domain-specific question-answer pairs; the latter are oversampled in order to better handle the imbalances. The results reveal considerable performance increases across all neural network architectures. The relative improvements can reach up to \SI{17.0}{\percent}.


The results clearly demonstrate that the fuse-and-oversample approach, attains consistently a superior performance. Its relative performance improvements range between \num{3.2} and \num{13.0} percentage points higher than for strategy~(ii). An explanation is that fine-tuning network parameters on a domain-specific dataset of such small size is a challenging undertaking, as one must manually calibrate the number of epochs, batch size and learning rate in order to avoid overfitting. For example, a batch size of \num{64} yields an entire training epoch on our dataset that consists of only six training steps. This in turn makes hyperparameter selection highly fragile. In contrast, training on the fused data proves to be substantially more robust and, in addition, requires less knowledge of training the network parameters.

\begin{table}[t]
	\small
	\centering
	\begin{tabular}{l cc cc cc}
		\toprule
		\mcellt{Neural network} & \multicolumn{2}{c}{\mcellt{Baseline:\\ no transfer learning}} & \multicolumn{2}{c}{\mcellt{Transfer learning:\\ init and fine-tune}} & \multicolumn{2}{c}{\mcellt{Transfer learning:\\ fuse-and-oversample}} \\
		\cmidrule(lr){2-3}\cmidrule(lr){4-5}\cmidrule(lr){6-7}
		& \multicolumn{1}{c}{EM} & \multicolumn{1}{c}{F1} & \multicolumn{1}{c}{EM} & \multicolumn{1}{c}{F1} & \multicolumn{1}{c}{EM} & \multicolumn{1}{c}{F1} \\
		\midrule
		DrQA          &  51.4  &  66.3  &  53.2   	 &  67.3   & \bfseries 59.9$^{**}$ & \bfseries 72.2   \\
		& 	 &        &\footnotesize  (\SI{+3.5}{\percent}) &  \footnotesize(\SI{+1.5}{\percent})     &  \footnotesize (\SI{+16.5}{\percent})     &  \footnotesize(\SI{+8.9}{\percent})     \\[0.3em]
		R-Net         &  38.9  & 55.2   & 41.2       &  56.5   & \bfseries 45.5$^{*}$ &  \bfseries  60.5 \\
		& 	 &        & \footnotesize(\SI{+5.9}{\percent})&\footnotesize  (\SI{+2.4}{\percent})    &  \footnotesize (\SI{+17.0}{\percent})     &  \footnotesize(\SI{+9.6}{\percent})    \\
		BiDAF 	 	  &  53.3  &  67.4  &  55.6$^{\dagger}$  	 &  68.4   & \bfseries 57.6$^{\dagger}$ &  \bfseries  70.0 \\[0.3em]
		& 	 &        &\footnotesize  (\SI{+4.3}{\percent})  & \footnotesize(\SI{+1.4}{\percent})     & \footnotesize  (\SI{+8.1}{\percent})   & \footnotesize(\SI{+3.9}{\percent})     \\
		\midrule
		\multicolumn{7}{l}{\footnotesize Significance levels: $^\dagger$ $0.1$; $^*$ $0.05$; $^{**}$ $0.01$; $^{***}$ $0.001$} \\
	\end{tabular}
	\caption{Sensitivity analysis comparing different methods for transfer learning. Here it is solely the accuracy of the answer extraction module that is evaluated; that is, the correct document is given and only the location of the correct answer is unknown. Accordingly, the performance achieves slightly higher values in comparison to earlier assessments of the overall system. Transfer learning based on a fused dataset yields a consistently superior performance as compared to the na{\"i}ve two-stage approach. The performance of each network architecture relative to its baseline is reported and the best-performing approach is highlighted in bold. We further performed McNemar's test in order to assess whether improvements in exact matches (EM) from transfer learning over the baseline are statistically significant.}
	\label{tab:results_answer_extraction}
\end{table}

\subsection{Domain Customization}


Finally, we evaluate the different approaches to domain customization within the entire Q\&A pipeline. The overall performance is measured by the fraction of \emph{exact matches}~(EM) with the ground-truth answer. Answers in the context of Q\&A are only counted as an exact match when the extracted candidate represents the shortest possible sub-span with the correct answer. Even though this is identical to accuracy, we avoid this term here in order to prevent misleading interpretations and emphasize the characteristics of the shortest sub-span. In addition, we measure the relative overlap between the candidate output and the shortest answer span by reporting the proportional match between both bag-of-words representations, yielding an macro-averaged F1-score for comparison. As a specific caveat, we follow common conventions and compute the metrics by ignoring punctuations and articles.


\Cref{tab:results_domain_customization} reports the numerical results. In addition to the three deep learning based models we use two baseline models as comparison. For implementation details, we refer to \Cref{sec:answer_extraction}. For all methodological strategies, we first list the performance without domain customization as a benchmark and, subsequently, incorporate the different approaches to domain customization. Without domain customization, the Q\&A system can (at best) answer one in \num{7.6} questions correctly, while the performance increases to one in \num{3.7} with the use of the additional metadata information. 
\begin{table}[t]
	\centering
	\small
	\sisetup{input-symbols={()\,\%}}
	\begin{tabular}{l cc cc cc cc}
		\toprule 	 	 	 	 	 
		Method & \multicolumn{2}{c}{\mcellt{No domain\\customization}} & \multicolumn{2}{c}{\mcellt{Fuse-and-oversample \\transfer learning}} & \multicolumn{2}{c}{\mcellt{Metadata\\filter}} & \multicolumn{2}{c}{\mcellt{Transfer learning\\ + metadata filter}} \\
		\cmidrule(lr){2-3}\cmidrule(lr){4-5}\cmidrule(lr){6-7}\cmidrule(lr){8-9}
		&\multicolumn{1}{c}{EM}&\multicolumn{1}{c}{F1}&\multicolumn{1}{c}{EM}&\multicolumn{1}{c}{F1}&\multicolumn{1}{c}{EM}&\multicolumn{1}{c}{F1}&\multicolumn{1}{c}{EM}&\multicolumn{1}{c}{F1}\\
		\midrule
		\multicolumn{9}{c}{\textsc{Baseline systems}} \\
		\midrule
		Sliding window         &5.4 &7.1&n/a&n/a&11.7&15.9&n/a&n/a\\[0.3em]
		Logistic regression    &13.2 &21.0&n/a&n/a&27.2&39.5&n/a&n/a\\
		\midrule
		\multicolumn{9}{c}{\textsc{Deep learning systems}} \\
		\midrule
		DrQA & 24.9 & 34.9 & 29.2 & 39.6 & 44.4 & 57.6 & \bfseries 51.0 &\bfseries 63.7 \\
		(best-fit document) &&&\small (\SI{+17.3}{\percent}) &\small (\SI{+13.5}{\percent})&\small (\SI{+78.3}{\percent})&\small (\SI{+65.0}{\percent})&\small (\SI{+104.8}{\percent})& \small (\SI{+82.5}{\percent})\\[0.3em]
		DrQA  &\bfseries 30.0 &\bfseries	41.6 & \bfseries 33.9 &\bfseries 45.4 & 40.1 & 52.4 & 47.1 & 59.3 \\
		(reference implementation)&&&\small (\SI{+13.0}{\percent}) &\small (\SI{+9.1}{\percent})&\small (\SI{+33.7}{\percent})&\small (\SI{+26.0}{\percent})&\small (\SI{+57.0}{\percent})& \small (\SI{+42.5}{\percent})\\[0.3em]
		R-Net      			  & 18.7  & 26.4 & 22.2 & 28.5 & 33.9 & 48.4 & 38.1 & 50.7 \\
		&&&\small (\SI{+18.7}{\percent}) &\small (\SI{+8.0}{\percent})&\small (\SI{+81.3}{\percent})&\small (\SI{+83.3}{\percent})&\small (\SI{+103.7}{\percent})& \small (\SI{+92.0}{\percent})\\[0.3em]
		BiDAF       		  & 26.5  &	33.0 & 28.4 & 33.8 & \bfseries 46.3 &\bfseries 58.7 & 49.8 & 60.6 \\
		&&&\small (\SI{+7.6}{\percent}) &\small (\SI{+2.4}{\percent})&\small (\SI{+75.4}{\percent})&\small (\SI{+77.9}{\percent})&\small (\SI{+88.6}{\percent})& \small (\SI{+83.6}{\percent})\\
		\bottomrule    
	\end{tabular}
	\centering
	\caption{Performance comparison of different strategies for domain customization. Here the plain system without domain customization is benchmarked against transfer learning and the additional selection mechanisms by firm name. Additionally, relative performance improvements over the baseline without domain customization are reported for each implementation with additional highlighting in bold for the best-performing system in each experimental setup. Notably, transfer learning is not applicable to the baseline systems.}
	\label{tab:results_domain_customization}
\end{table}


All algorithms incorporating deep learning clearly outperform the baselines from the literature. For instance, the DrQA system yields an exact match in one out of \num{3.3} cases. Here we note that two DrQA systems are compared: namely, one following the reference implementation \cite{Chen.2017} -- whereby five documents are returned by the information retrieval module and the extracted answer is scored for each -- and, for reasons of comparability, one approach that is based on the best-fit document, analogous to the other neural-network-based systems. 


We observe considerable performance improvements as a result of applying fuse-and-oversample transfer learning. It can alone increase the ratio of exact matches to one out of \num{2.4} questions and, together with the metadata filtering, it even achieves a score of one out of \num{2.2}. This increases the exact matches of the best-scoring benchmark without domain customization by \num{3.9} and \num{19.8}, respectively. The gains yielded by the selection mechanism come naturally, but we observe that a limited, domain-specific training set can also boost the performance considerably. Notably, neither of the baseline systems can be further improved with transfer learning, as this technique is not applicable (\eg the sliding window approach lacks trainable parameters).


The DrQA system that considers the top-five documents yields the superior performance in the first two experiments: the plain case and the one utilizing transfer learning. However, the inclusion of the additional selection mechanism alters the picture, as it essentially eliminates the benefits of returning more than one document in the information retrieval module, an effect explained in~\cite{Kratzwald.2018c}. Apparently, extracting the answer from multiple documents is useful in settings where the information retrieval module is less accurate, whereas in settings with precise information retrieval modules, the additionally returned documents introduce unwanted noise and prove counterproductive. The result also goes hand in hand with our finding that, due to the additional metadata filter, the information retrieval module becomes very accurate and, as a result, the performance of the answer extraction module becomes especially critical, since it usually represents the most sensitive part of the Q\&A pipeline.

\begin{table}[t]
	\small
	\centering
	\begin{tabular}{l cc cc cc}
		\toprule
		\mcellt{Neural network} & \multicolumn{2}{c}{\mcellt{Baseline:\\ no transfer learning}} & \multicolumn{2}{c}{\mcellt{Transfer learning:\\ init and fine-tune}} & \multicolumn{2}{c}{\mcellt{Transfer learning:\\ fuse-and-oversample}} \\
		\cmidrule(lr){2-3}\cmidrule(lr){4-5}\cmidrule(lr){6-7}
		& \multicolumn{1}{c}{EM} & \multicolumn{1}{c}{F1} & \multicolumn{1}{c}{EM} & \multicolumn{1}{c}{F1} & \multicolumn{1}{c}{EM} & \multicolumn{1}{c}{F1} \\
		\midrule
		DrQA          &  48.2 & 55.2 & 54.2$^{**}$ & 60.4 &	\bfseries56.8$^{***}$	&	\bfseries62.5  \\
		& 	 &        & \footnotesize (\SI{12.4}{\percent}) &\footnotesize  (\SI{9.4}{\percent})     &\footnotesize   (\SI{17.8}{\percent})     &  \footnotesize (\SI{13.2}{\percent})     \\[0.3em]
		R-Net         &  43.4 & 48.9 & 47.6$^{*}$ & 53.5 & \bfseries 49.0$^{**}$ &\bfseries56.2\\
		& 	 &        & \footnotesize(\SI{+9.7}{\percent})& \footnotesize (\SI{9.4}{\percent})    &\footnotesize   (\SI{+12.9}{\percent})     &\footnotesize  (\SI{14.9}{\percent})    \\
		BiDAF 	 	  &  48.6 & 53.6 & 52.2$^{\dagger}$ & 56.4 & \bfseries58.6$^{**}$ & \bfseries62.2 \\[0.3em]
		& 	 &        & \footnotesize (\SI{7.4}{\percent})  &\footnotesize (\SI{5.2}{\percent})     & \footnotesize  (\SI{20.5}{\percent})   &\footnotesize (\SI{16.0}{\percent})     \\
		\midrule
		\multicolumn{7}{l}{\footnotesize Significance levels: $^\dagger$ $0.1$; $^*$ $0.05$; $^{**}$ $0.01$; $^{***}$ $0.001$}\\
	\end{tabular}
	\caption{Robustness analysis of the different transfer learning techniques on a different dataset. Transfer learning based on a fused dataset yields a consistently superior performance as compared to the na{\"i}ve two-stage approach. The performance of each network architecture relative to its baseline is reported and the best-performing approach is highlighted in bold. We further performed McNemar's test in order to assess whether improvements in exact matches (EM) from transfer learning over the baseline are statistically significant.}
	\label{tab:results_answer_extraction2}
\end{table}

\subsection{Robustness Check: Transfer Learning with Additional Domain Customization}

To further demonstrate the robustness of our transfer learning approach, we compiled a second dataset for a use case from the movie industry. Therefore, we randomly subsampled 393 training and 257 test question-answer pairs from the WikiMovies dataset~\cite{Miller.2016}, and annotated them with a supporting text passage from Wikipedia. Here we restrict our analysis to the transfer learning approach as metadata filtering can be transferred in a straightforward manner, whereas domain customization of the answer extraction component presents the more challenging and fragile element with regard to performance. We finally run a robustness check on a second dataset containing movie questions. The results shown in \Cref{tab:results_answer_extraction2}, confirm our previous findings where the fuse-and-oversample approach consistently outperforms the na\"ive method from literature. Our holistic evaluation over different architectures and datasets contributes to the generalizability of our results.

\section{Discussion}
\label{sec:discussion}

\subsection{Domain Customization}


Hitherto, a key barrier to the widespread use of Q\&A systems has been the inadequate accuracy of such systems. Challenges arise especially when practical applications, such as those in the domain of finance, entail complex language with highly special terminology. This requires an efficient strategy for customizing Q\&A systems to domain-specific characteristics. Our paper proposes and evaluates two such levers: incorporating metadata for choosing sub-domains and transfer learning. Both entail fairly small upfront costs and generalize across all domains and application areas, thereby ensuring straightforward implementation in practice. 


Our results demonstrate that domain customization greatly improves the performance of question-answering systems. Metadata is used for sub-domain filtering and increases the number of exact matches with the shortest correct answer by up to \SI{81.3}{\percent}, while transfer learning yields gains of up to \SI{18.7}{\percent}. The use of metadata and transfer learning presents an intriguing, cost-efficient path to domain customization, that has so far been overlooked in systems working on top of unstructured text. Transfer learning enables an inductive transfer of knowledge in the Q\&A system from a different, unrelated dataset to the domain-specific application, for which one can utilize existing datasets that sometimes include more than \num{100000} entries and are publicly available.  


\subsection{Design Challenges}

During our research, we identified multiple challenges in customizing QA systems to domain-specific applications. These are summarized in the following together with possible remedies. 
\begin{description}
	\item[Labeling effort.]  The manual process of labeling thousands of question-answer pairs for each domain-specific application renders deep neural networks unfeasible. With the use of transfer learning, only a few hundred samples are sufficient for training the deep neural networks and achieving significant performance improvements. In our case, the system requires a small dataset of as few as \num{400} annotated question-answer pairs. This is especially beneficial in business settings where annotations demand extensive prior knowledge (such as in medicine or law) since here the necessary input from domain experts is greatly reduced.
	\item[Multi-component architecture.] The accuracy of the information retrieval module is crucial since it upper-bounds the end-to-end performance of our system; \ie, we cannot answer questions for which we cannot locate the document containing the answer. While the choice of the actual retrieval model seems to be less important, we observe a strong improvement after incorporating metadata information. Hence, practitioners want to consider this trade-off when making strategic choices regarding the implementation of Q\&A systems.
	\item[Hyperparameter tuning.] Traditional initialize-and-fine-tune transfer learning proves to be non-trivial. Due to the limited amount of data, the correct selection of the learning rate and other hyper-parameters becomes extremely volatile. As remedy, we proposed a transfer learning approach on fuse-and-oversample which allows to achieve better performance and renders hyperparameter tuning unnecessary. 
	\item[Design of metadata filtering.] This paper demonstrates the potential of metadata filtering as a means for customizing Q\&A systems to domain-specific applications. Despite the prevalence of metadata in information systems, its use has been largely overlooked for question answering over unstructured content. In our use case, we achieve remarkable performance improvements already from a fairly na{\"i}ve metadata filter. However, the actual design must be carefully adapted by practitioners to the domain-specific use case. In future research, a promising path would be to study advanced filtering mechanisms that extend to open-domain settings and incorporate additional semantic information, as well as to compare different design choices from a behavioral perspective. 
\end{description}

We apply our approach to examples from two different domains, while using three different information retrieval modules and up to five answer extraction modules. In general, our approach is extensible to other domains in a straightforward manner. The only assumptions we make are common in prior research \cite{Chen.2017,Wang.2018} and, hence, point towards the boundaries of our artifact. First, the language is limited to English, as our large-scale corpus for transfer learning is in English and similar datasets for other languages are still rare. Second, answers are required to be a sub-span of a single document; however, this is currently a limiting factor in almost all content-based Q\&A systems. Third, questions must be answerable with the information that is contained in a single document. These assumptions generally hold true for factoid question-answering tasks \cite{Voorhees.2001} across multiple disciplines. Fourth, we assume that a meaningful filter for metadata can be derived in domain-specific applications; however, its benefit can vary depending on the actual nature of the metadata.

\subsection{Implications for Research}


Our experiments reveal another source of performance improvements beyond domain customization: replacing the known answer extraction modules with transfer learning and, instead, tapping advanced neural network architectures into the Q\&A system. This is interesting in light of the fact that deep neural networks have fostered innovations in various areas of natural language processing, and yet publications pertaining to question answering are limited to a few exceptions \cite[\cf][]{Chen.2017, Wang.2017.b, Wang.2018}. Additional advances in the field of deep learning are likely to present a path towards further bolstering the accuracy of the system. The great improvements achieved by including metadata promises an interesting path for future research that could be generalized and automated for open-domain applications.


Future research could evolve Q\&A systems in several directions. First, considerable effort will be needed to overcome the current approach whereby answers can only be sub-spans of the original documents and, instead, devise a method in which the system re-formulates the answer by combining information across different documents. Second, Q\&A systems should be extended to better handle semi-structured information such as tables or linked data, since these are common in today's information systems. Third, practical implementations could benefit from ensemble learning \cite[\cf][]{Wang.2017,Seo.2016}. 

\subsection{Conclusion}
\label{sec:conclusion}


Users, and especially corporations, demand efficient access to the knowledge stored in their information systems in order to fully inform their decision-making. Such retrieval of information can be achieved through question-answering functionality. This can overcome limitations inherent to the traditional keyword-based search. More specifically, Q\&A systems are known for their ease of use, since they enable users to interact conveniently in natural language. This also increases the acceptance rates of information systems in general and accelerates the overall search process. Despite these obvious advantages, inefficient domain customization represents a major barrier to Q\&A usage in real-world applications, a problem for which this paper presents a powerful remedy.  


This work contributes to the domain customization of Q\&A systems. We first demonstrate that practical use cases can benefit from including metadata information during information retrieval. Furthermore, we propose the use of transfer learning and specifically our novel fuse-and-oversample approach in order to reduce the need for pre-labeled datasets. Only relatively small sets of question-answer pairs are needed to fine-tune the neural networks, whereas the majority of the learning process occurs through an inductive transfer of knowledge. Altogether, this circumvents the needs for hand-labeling thousands of question-answer pairs as part of tailoring question answering to specific domains; instead, the proposed methodology requires comparatively little effort and thus allows even small businesses to take advantage of deep learning.

\appendix

\section{Appendix}
\label{sec:appendix}

\subsection{Information Retrieval Module}
\label{sec:information_retrieval}

The information retrieval module is responsible for locating documents relevant to the given question. In this work, we implemented three different modules: a vector space model based on cosine similarity scoring, the probabilistic Okapi BM25 model and a tf-idf model based on hashed bigram counts as follows.

\textbf{Vector Space Model:}  Let $\mathit{tf}_{ji}$ refer to the term frequencies of document $i = 1, \ldots, N$ for vocabulary $j = 1, \ldots T$. In order to better identify characteristic terms, the term frequencies are weighted by the inverse document frequency, \ie giving the tf-idf score $w_{ji} = \mathit{tf}_{ji} \, \mathit{idf}_j$ \cite{SparckJones.1972}. Here the inverse document frequency places additional discriminatory power on terms that appear only in a subset of the documents. It is defined by $\mathit{idf}_j = \log( N / n_j)$ where $n_j$ denotes the number of documents that entail the term $j$. This translates a document $i$ into a vector representation $d_i = \left[ w_{1i}, w_{2i}, \ldots, w_{Ti} \right]^T$. Analogously, queries are also processed to yield a vector representation $q$. The relevance of a document $d_i$ to a question $q$ can then be computed by measuring the cosine similarity between both vectors. This is formalized by 
\begin{equation}
\cos(d_i,q) = \frac{d_i^T q}{\Vert d_i\Vert \, \Vert q\Vert}.
\end{equation} 
Subsequently, the information retrieval module determines the document $d_q = \argmin_{d_i} \cos(d_i,q)$ that displays the greatest similarity between document and question. 

\textbf{Okapi BM25 Model:} The Okapi BM25 refers to family of probabilistic retrieval models that provide state-of-the-art performance in plain document and information retrieval. Rather than using vectors to represent documents and queries, this model uses a probabilistic approach to determine the probability of a document given a query. We used the default implementation\footnote{Provided by gensim \url{https://radimrehurek.com/gensim/summarization/bm25.html}} as described in \cite{Robertson.2009}.

\textbf{Bigram Model: } This approach yields state-of-the-art results in many recent applications. The tf-idf weighting in our vector space model ignores semantics, such as the local ordering of words, and, as a remedy, we incorporate $n$-grams instead. In order to deal with high number of possible $n$-grams and, therefore, the high dimensionality of tf-idf vectors, we utilize feature hashing~\cite{Weinberger.2009}. We stick with prior work using bi-grams and construct the tf-idf vectors in the same way as described in \cite{Chen.2017, Wang.2018}. Finally, the score of a document is given by the dot-product between query and document vector.

\subsection{Answer Extraction Module}
\label{sec:answer_extraction}

In the second stage, the answer extraction module draws upon the previously-selected document and extracts the answer $a \in d_q$. Based on our literature review, this work evaluates different baselines for reasons of comparability, namely, two benchmarks utilizing traditional machine learning and the DrQA network from the field of deep learning. Furthermore, we suggest the use of two additional deep neural networks that advance the architecture beyond DrQA. More precisely, our networks incorporate character-level embeddings and an interplay of different attention mechanisms, which together allow us to better adapt to unseen words and the context of the question. 

\subsubsection{Baseline Methods}


We implement two baselines from previous literature, namely, a sliding window approach without trainable parameters \cite{Richardson.2013} and a machine learning classifier based on lexical features \cite{Rajpurkar.2016}. Both extract linguistic constituents from the source document to narrow down the number of candidate answers. Here the concept of a constituent refers to one or multiple words that can stand on their own (\eg nouns, a subject or object, a main clause).

The sliding window approach processes the text passage and chooses the sub-span of words as an answer that has the highest number of overlapping terms with the question. The second approach draws upon a logistic regression in order to rank candidate answers based on an extensive series of hand-crafted lexical features. The choice of features contains, for instance, tf-idf weights extended with lexical information. We refer to \cite{Rajpurkar.2016} for a description of the complete list. The classifier is subsequently calibrated using a training set of documents and correct responses in order to select answers for unseen question-answer pairs. 

\subsubsection{Deep Learning Methods}


Prior work \cite{Chen.2017} has proposed the use of deep learning within the answer extraction module, resulting in the DrQA network, which we utilize as part of our experiments. Furthermore, we draw upon additional network architectures, namely, BiDAF \cite{Seo.2016} and R-Net~\cite{Wang.2017}, which were recently developed for the related, yet different, task of machine comprehension.\footnote{The task of machine comprehension refers to locating text passages in a given document and thus differs from question answering, which includes the additional search as part of the information retrieval module. These models have shown significant success recently; yet they require the text passage containing the answer to be known up front.} Accordingly, we modify two state-of-the-art machine comprehension models such that they work within our Q\&A pipeline. These network architectures incorporate character-level embeddings which allow for the handling of unseen vocabulary and, in practice, find more suitable numerical representations for infrequent words. Second, the attention mechanism is modeled in such a way that it simultaneously incorporates both question and answer, which introduces additional degrees-of-freedom for the network, especially in order to weigh responses such that the context matches. 

\begin{figure}[t]
	\begin{center}
		\includegraphics[width=.8\linewidth]{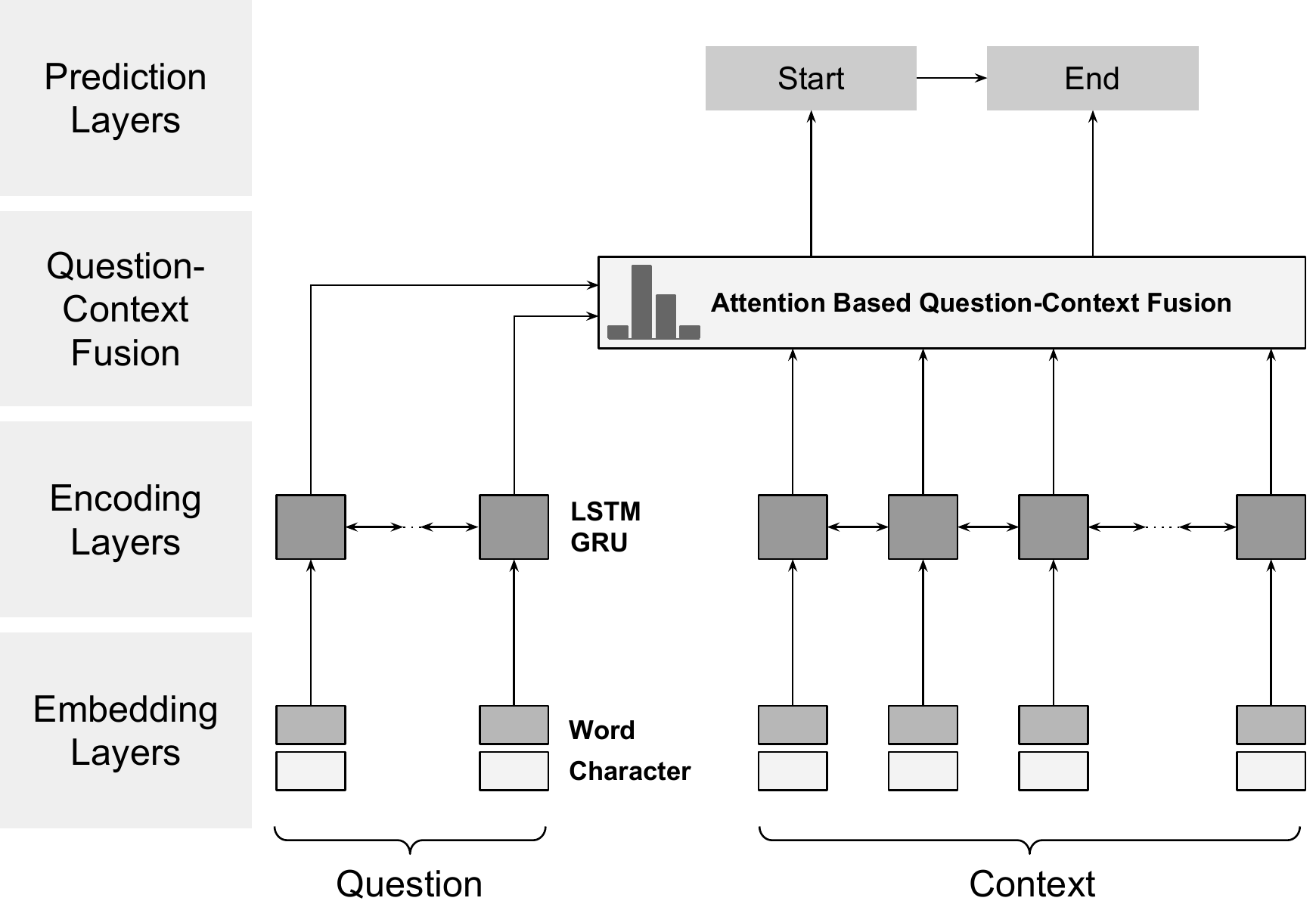}
	\end{center}
	\caption{Schematic architecture of the different neural network architectures (\ie DrQA, BiDAF, R-Net) used in the answer extraction module.}
	\label{fig:rc_arch}
\end{figure}


In the following, we summarize the key elements of the different neural networks. The architectures entail several differences across the networks, but generally follow the schematic structure in \Cref{fig:rc_arch} consisting of embedding layers, encodings through recurrent layers, an attention mechanism for question-answer fusion and the final layer predicting both the start and end position of the answer.   


\textbf{Embedding layers.} The first layer in neural machine comprehension networks is an embedding layer whose purpose is the replacement of high-dimensional one-hot vectors that represent words with low-dimensional (but dense) vectors. These vectors are embedded in a semantically meaningful way in order to preserve their contextual similarity. Here all networks utilize the word-level embeddings yielded by \emph{glove}~\cite{Pennington.2014}.  For both R-Net and BiDAF, additional character-level embeddings are trained to complement the word embeddings for out-of-dictionary words. At the same time, character-level embedding can still yield meaningful embeddings even for rare words with which embeddings at word level struggle due to the small number of samples. Differences between R-Net and BiDAF arise with regard to the way in which character- and word-level embeddings are fed into the next layer. R-Net computes a simple concatenation of both vectors, while BiDAF fuses them with an additional two-layer highway network.


\textbf{Encoding layers.} The output from the embedding layer for the question and context are fed into a set of recurrent layers. Recurrent layers offer the benefit of explicitly modeling sequential structure and thus encode a complete sequence of words into a fixed-size vector. Formally, the output $o_j$ of a recurrent layer when processing the $j$-th term is calculated from the $j$-th hidden state via $o_j = f(h_j)$. The hidden state is, in turn, computed from the current input $x_j$ and the previous hidden state $h_{j-1}$ via $o_j = g(h_{j-1}, x_j)$, thereby introducing a recurrent relationship. The actual implementation of $f(\ldots)$ and $g(\ldots)$ depends on the architectural choice: BiDAF and DrQA utilize long short-term memories, while R-Net instead draws upon gated recurrent units, which are computationally cheaper but also offer less flexibility. All models further extend these networks via a bidirectional structure in which two recurrent networks process the input from either direction simultaneously. 


\textbf{Question-context fusion.} Both questions and context have been previously been processed separately and these are now combined in a single mathematical representation. To facilitate this, neural networks commonly employ an attention mechanism~\cite{Bahdanau.2014}, which introduces an additional set of trainable parameters in order to better discriminate among individual text segments according to their relevance in the given context. As an example, the interrogative pronoun \emph{"who"} in a question suggests that the name of a person or entity is sought and, as a result, the network should focus more attention on named entities. Mathematically, this is achieved by an additional dot product between the embedding of \emph{"who"} and the words from the context, which is further parametrized through a softmax layer. The different networks vary in in how they implement the attention mechanisms. The DrQA draws upon a fairly simple attention mechanism, while both BiDAF and R-Net utilize a combination of multiple attention mechanisms. 


\textbf{Prediction layer.} The final prediction is responsible for determining the beginning and ending position of the answer within the context. DrQA utilizes two independent classifiers for making the predictions. This has the potential disadvantage that the ending position does not necessarily come after the starting position. This is addressed by both BiDAF and R-Net, where the prediction of the end position is conditioned on the predicted beginning. Here the BiDAF network simply combines the outputs of the previous layers in order to make the predictions, while R-Net implements an additional pointer network. 

\begin{acks}
	Cloud computing resources were provided by a Microsoft Azure for Research award. We appreciate the help of Ryan Grabowski in editing our manuscript with regard to language. 
\end{acks}

\bibliographystyle{model5-names}
\bibliography{qa_news}

\end{document}